\documentclass[11pt,a4paper]{article}
\usepackage[utf8]{inputenc}
\usepackage[hyperref]{acl2018}
\usepackage{times}
\usepackage{latexsym}
\usepackage{color}
\usepackage{listings}
\usepackage{url}
\usepackage{cleveref}
\usepackage{textcomp}
\usepackage{caption}
\usepackage{mathtools}
\usepackage{wrapfig}
\usepackage{dblfloatfix}

\aclfinalcopy 
\def\aclpaperid{69} 

\let\tempone\itemize
\let\temptwo\enditemize
\renewenvironment{itemize}{\tempone\setlength{\topsep}{0pt}\setlength{\partopsep}{0pt}\setlength{\itemsep}{0pt}}{\temptwo}

\usepackage{enumitem}
\setlist{
	itemsep=0pt,
	parsep=0pt plus 2pt minus 1pt,
	topsep=0pt plus 2pt minus 1pt,
	partopsep=0pt}

\definecolor{deepblue}{rgb}{0,0,0.5}
\definecolor{deepred}{rgb}{0.6,0,0}
\definecolor{deepgreen}{rgb}{0,0.4,0}
\definecolor{gray}{rgb}{0.5,0.5,0.5}

\newcommand{\crnn}{RETURNN} 

\title{\crnn{} as a Generic Flexible Neural Toolkit with Application to Translation and Speech Recognition}

\author{{
\bf Albert Zeyer$^{1,2,3}$, Tamer Alkhouli$^{1,2}$
and Hermann Ney$^{1,2}$
} \\
$^1$Human Language Technology and Pattern Recognition Group \\
RWTH Aachen University, Aachen, Germany, \\
$^2$AppTek, USA, \url{http://www.apptek.com/}, \\
$^3$NNAISENSE, Switzerland, \url{https://nnaisense.com/} \\
{\tt {surname}@cs.rwth-aachen.de}
}

\date{}

\begin{document}
\maketitle
\begin{abstract}
We compare the fast training and decoding speed of \crnn{}
of attention models for translation,
due to fast CUDA LSTM kernels,
and a fast pure TensorFlow beam search decoder.
We show that a layer-wise pretraining scheme
for recurrent attention models
gives over 1\% BLEU improvement absolute
and it allows to train deeper recurrent encoder networks.
Promising preliminary results on max.\ expected BLEU training are presented.
We obtain state-of-the-art models trained on the  WMT 2017 German$\leftrightarrow$English 
translation task. We also present end-to-end model results for speech recognition on the Switchboard
task.
The flexibility of \crnn{} allows a fast research feedback loop
to experiment with alternative architectures,
and its generality allows to use it on a wide range of applications.
\end{abstract}

\section{Introduction}

\crnn{}, the RWTH extensible training framework for universal recurrent neural networks,
was introduced in \cite{doetsch2017:returnn}.
The source code is fully open\footnote{\tiny\url{https://github.com/rwth-i6/returnn}}.
It can use Theano \cite{theano2016} or TensorFlow \cite{tensorflow2015} for its computation.
Since it was introduced, it got extended by
comprehensive TensorFlow support.
A generic recurrent layer
allows for a wide range of encoder-decoder-attention
or other recurrent structures.
An automatic optimization logic can
optimize the computation graph depending on
training, scheduled sampling, sequence training,
or beam search decoding.
The automatic optimization together with our fast
native CUDA implemented LSTM kernels
allows for very fast training and decoding.
We will show in speed comparisons with
 Sockeye \cite{hieber2017sockeye}
that we are at least as fast or usually faster
in both training and decoding.
Additionally, we show in experiments that we can
train very competitive models for machine translation
and speech recognition.
This flexibility together with the speed
is the biggest strength of \crnn{}.

Our focus will be
on recurrent attention models.
We introduce a layer-wise pretraining scheme
for attention models and show its significant effect
on deep recurrent encoder models.
We show promising preliminary results on expected maximum BLEU training.
The configuration files of all the experiments are publicly available%
\footnote{\tiny\url{https://github.com/rwth-i6/returnn-experiments/tree/master/2018-attention}}.

\section{Related work}

Multiple frameworks exist for training attention models,
most of which are
focused on machine translation.
\begin{itemize}
\item
Sockeye \cite{hieber2017sockeye} is a generic framework
based on MXNet \cite{chen2015mxnet}
which is most compareable to \crnn{} as it is generic
although we argue that \crnn{} is more flexible and faster.
\item
OpenNMT \cite{levin2017opennmt1,levin2017opennmt2}
based on Lua \cite{Ierusalimschy:2006:LRM:1215067}
which is discontinued in development.
Separate PyTorch \cite{pytorch2018} and TensorFlow implementation
exists, which are more recent.
We will demonstrate that \crnn{} is more flexible.
\item
Nematus \cite{sennrich2017nematus}
is based on Theano \cite{theano2016}
which is going to be discontinued in development.
We show that \crnn{} is much faster in both training and decoding
as can be concluded from our speed comparison to Sockeye
and the
comparisons performed by the Sockeye authors \cite{hieber2017sockeye}.
\item
Marian \cite{junczys2016marian} is implemented directly in C++ for performance reasons.
Again by our speed comparisons
and the comparisons performed by the Sockeye authors \cite{hieber2017sockeye},
one can conclude
that \crnn{} is very competitive in terms of speed,
but is much more flexible.
\item
NeuralMonkey \cite{helcl2017neuralmonkey}
is based on TensorFlow \cite{tensorflow2015}.
This framework is not as flexible as \crnn{}.
Also here we can conclude just as before
that \crnn{} is much faster in both training and decoding.
\item
Tensor2Tensor \cite{vaswani2018tensor2tensor}
is based on TensorFlow \cite{tensorflow2015}.
It comes with the reference implementation of the Transformer model \cite{vaswani2017ffatt},
however, it lacks support for recurrent decoder models
and overall is way less flexible than \crnn{}.

\end{itemize}

\section{Speed comparison}

Various improved and fast CUDA LSTM kernels are available
for the TensorFlow backend in \crnn{}.
A comparison of the speed of its own LSTM kernel vs.\ other TensorFlow LSTM kernels
can be found on the website\footnote{\tiny\url{http://returnn.readthedocs.io/en/latest/tf_lstm_benchmark.html}}.
In addition, an automatic optimization path which moves out computation
of the recurrent loop as much as possible improves the performance.

We want to compare different toolkits in training and decoding
for a recurrent attention model in terms of speed on a GPU.
Here, we try to maximize the batch size
such that it still fits into the GPU memory of our reference GPU card,
the Nvidia GTX 1080 Ti with 11 GB of memory.
We keep the maximum sequence length in a batch the same,
which is 60 words.
We always use Adam \cite{kingma2014adam} for training.
In \Cref{tab:trainspeed}, we see that \crnn{} is the fastest,
and also is most efficient in its memory consumption (implied by the larger batches).
For these speed experiments, we did not tune any of the hyper parameters
of \crnn{} which explains its worse performance. The aim here is 
to match Sockeye's exact architecture for speed and memory comparison.
During training, we observed that the learning rate scheduling settings
of Sockeye are more pessimistic, i.e.\ the decrease is slower
and it sees the data more often until convergence.
This greatly increases the total training time but in our experience
also improves the model.

\begin{table}[t]
\setlength{\tabcolsep}{0.3em}
\centerline{
\begin{tabular}{|l|c|c|c|c|c|}
\hline
  toolkit &  encoder &  time  &  batch &  \multicolumn{2}{c|}{BLEU [\%]} \\
         & n. layers &       [h]    &   size  & 2015 & 2017 \\
  \hline \hline
	\crnn{} & 4 & \textbf{11.25} & 8500 & 28.0 & 28.4 \\ \cline{1-1} \cline{3-6}
	Sockeye & & 11.45 & 3000 & \textbf{28.9} & \textbf{29.2} \\ \cline{1-1} \cline{3-6}
 \hline \hline
	\crnn{} & 6 & \textbf{12.87} & 7500 & 28.7 & 28.7  \\ \cline{1-1}\cline{3-6}
	Sockeye & & 14.76 & 2500 & \textbf{29.4} & \textbf{29.1} \\
\hline
\end{tabular}}
  \caption{Training speed and memory consumption on WMT 
  2017 German$\to$English.
  Train time is for seeing the full train dataset once.
  Batch size is in words, such that it almost  maximizes
  the GPU memory consumption.
  The BLEU score is for the  converged models,
  reported for newstest2015 (dev) and newstest2017.
  The encoder has one bidirectional LSTM layer and
  either 3 or 5 unidirectional LSTM layers.
  }
  \label{tab:trainspeed}
\end{table}

For decoding, we extend \crnn{}  with a fast pure TensorFlow beam search decoder,
which supports batch decoding and can run on the GPU.
A speed and memory consumption comparison is shown in \Cref{tab:decspeed}.
We see that \crnn{} is the fastest. We report results for the
batch size that yields the best speed. The slow
speed of Sockeye is due to frequent cross-device communication.

\begin{table}[th]
\setlength{\tabcolsep}{0.3em}
\centerline{
\begin{tabular}{|l|c|c|c|c|}
\hline
  toolkit &  encoder &  batch size &  \multicolumn{2}{c|}{time [secs]} \\
         & n. layers     &   [seqs]  & 2015 & 2017 \\
  \hline \hline
	\crnn{} & 4 & 50 & \textbf{54} & \textbf{71} \\ \cline{1-1} \cline{3-5}
	Sockeye & & 5 & 398 & 581 \\ \cline{1-1} \cline{3-5}
 \hline \hline
	\crnn{} & 6 & 50 & \textbf{56} & \textbf{70} \\ \cline{1-1}\cline{3-5}
	Sockeye & & 5 & 403 & 585 \\
\hline
\end{tabular}}
  \caption{Decoding speed and memory consumption
  on WMT 2017 German$\to$English.
  Time is for decoding the whole dataset,
  reported for  newstest2015 (dev) and newstest2017,
  with beam size 12.
  Batch size is the number of sequences,
  such that it optimizes the decoding speed.
  This does not mean that it uses the whole GPU memory.
  These are the same models as in \Cref{tab:trainspeed}.
  }
  \label{tab:decspeed}
\end{table}

\section{Performance comparison}

We want to study what possible performance we can get
with each framework on a specific task.
We restrict this comparison here to recurrent attention models.

The first task is the  WMT 2017 German to English translation task.
We use the same 20K byte-pair encoding subword units in all toolkits \cite{sennrich2015bpesubwords}.
We also use Adam \cite{kingma2014adam} in all cases.
The learning rate scheduling is also similar.
In \crnn{}, we use a 6 layer bidirectional encoder,
trained with pretraining and label smoothing.
It has bidirectional LSTMs in every layer of the encoder,
unlike Sockeye, which only has the first layer bidirectional.
We use a variant of attention weight / fertility feedback \cite{tu2016ACL},
which is inverse in our case, to use a multiplication instead of a division,
for better numerical stability.
Our model was derived from the model
presented by \cite{bahar17iwslt,peter17:wmt}
and \cite{bahdanau2014nmt}.

We report the best performing Sockeye model we trained, which
has 1 bidirectional and 3 unidirectional encoder layers, 1 
pre-attention target recurrent layer, and 1 post-attention
decoder layer. We trained with a max sequence length of 75,
and used the `coverage' RNN attention type. For Sockeye, the final
model is an average of the 4 best runs according to
the development perplexity.
The results are collected in \Cref{tab:perfwmt2017}.
We obtain the best results with Sockeye using a
Transformer network model \cite{vaswani2017ffatt}, where
we achieve 32.0\% BLEU on newstest2017.
So far, \crnn{} does not support this architecture;
see \Cref{sec:features} for details.

\begin{table}[th]
\setlength{\tabcolsep}{0.3em}
\centerline{
\begin{tabular}{|l|c|c|}
\hline
  toolkit &  \multicolumn{2}{c|}{BLEU [\%]} \\
            & 2015 & 2017 \\
 \hline \hline
	\crnn{} & \textbf{31.2} & \textbf{31.3} \\ \hline
	Sockeye & 29.7 & 30.2 \\
\hline
\end{tabular}}
  \caption{Comparison on German$\rightarrow$English.}
  \label{tab:perfwmt2017}
\end{table}

We compare \crnn{} to other toolkits on the WMT 2017 English$\to$German
 translation task in
\Cref{tab:perfwmt2017-ende}. We observe that our toolkit
outperforms all other toolkits. The best result obtained by other
toolkits is using Marian (25.5\% BLEU). In comparison, \crnn{} 
achieves 26.1\%.
We also compare \crnn{} to the best performing single systems of WMT 2017.
In comparison to the fine-tuned evaluation
systems that also include back-translated data, our model performs worse by only
0.3 to 0.9 BLEU. We did not run experiments with back-translated data, which can 
potentially boost the performance by several BLEU points.

\begin{table}[!th]
\setlength{\tabcolsep}{0.3em}
\centerline{
\begin{tabular}{|l|c|c|}
\hline
  System &  \multicolumn{1}{c|}{BLEU [\%]} \\
            &  newstest2017 \\
 \hline \hline
	\crnn{} 	& \textbf{26.1} \\ \hline 
	OpenNMT-py	& 21.8 \\
	OpenNMT-lua	& 22.6	\\
	Marian		& 25.6 \\
	Nematus		& 23.5 \\ 
	Sockeye		& 25.3 \\ \hline
	\multicolumn{2}{|c|}{WMT 2017 Single Systems + bt data } \\  \hline
	LMU 		& 26.4  \\
	 + reranking 	& 27.0 \\
	Systran 	& 26.5 \\
	Edinburgh 	& 26.5 \\
\hline
\end{tabular}}
	\caption{Performance comparison on WMT 2017 English$\rightarrow$German. The baseline systems (upper half) are trained
	  on the parallel data of the WMT Enlgish$\rightarrow$German 2017 task. We downloaded the hypotheses 
	  from here.\footnotemark \, The WMT 2017 system hypotheses (lower half) are generated using systems having additional back-
	  translation (bt)
	  data. These hypotheses are downloaded from here.\footnotemark}

  \label{tab:perfwmt2017-ende}
\end{table}


We also have preliminary results with recurrent attention models
for speech recognition on the Switchboard task, which we trained on the 300h trainset.
We report on both the Switchboard (SWB) and the CallHome (CH) part of Hub5’00 and Hub5’01.
We also compare to a conventional frame-wise trained
hybrid deep bidirectional LSTM with 6 layers \cite{zeyer17:lstm},
and a generalized full-sum sequence trained
hybrid deep bidirectional LSTM with 5 layers \cite{zeyer2017:ctc}.
The frame-wise trained hybrid model also uses focal loss \cite{lin2017focalloss}.
All the hybrid models use a phonetic lexicon and an external 4-gram language model
which was trained on the transcripts of both the Switchboard and the Fisher corpus.
The attention model does not use any external language model nor a phonetic lexicon.
Its output labels are byte-pair encoded subword units \cite{sennrich2015bpesubwords}.
It has a 6 layer bidirectional encoder,
which also applies max-pooling in the time dimension,
i.e.\ it reduces the input sequence by factor 8.
Pretraining as explained in \Cref{sec:pretrain} was applied.
To our knowledge, this is the best reported result for an end-to-end system
on Switchboard 300h without using a language model or the lexicon.
For comparison, we also selected comparable results from the literature.
From these, the Baidu DeepSpeech CTC model is modeled on characters
and does not use the lexicon but it does use a language model.
The results are collected in \Cref{tab:perfswb}.
	  \footnotetext[4]{ \url{https://github.com/awslabs/sockeye/tree/arxiv_1217/arxiv/output/rnn}}
	  \footnotetext[5]{ \url{http://matrix.statmt.org/}}
\begin{table}[th]
\setlength{\tabcolsep}{0.2em}
\centerline{
\begin{tabular}{|c|c|c|c|c|c|}
\hline
  model & training &  \multicolumn{4}{c|}{WER [\%]} \\
  & & \multicolumn{3}{c|}{Hub5'00} & Hub5'01 \\
           &    & $\Sigma$ & SWB & CH &  \\
 \hline \hline
 hybrid$^1$ & frame-wise & & 11.2 & & \\ \hline\hline
 hybrid$^2$ & LF-MMI & 15.8 & 10.8 & & \\ \hline\hline
 CTC$^3$ & CTC & 25.9 & 20.0 & 31.8 & \\
 \hline\hline\hline
 hybrid & frame-wise & \textbf{14.4} & \textbf{9.8} & \textbf{19.0} & 14.7 \\ \cline{2-6}
          & full-sum & 15.9 & 10.1 & 21.8 & \textbf{14.5} \\ \hline\hline
 attention & frame-wise & 20.3 & 13.5 & 27.1 & 19.9 \\
\hline
\end{tabular}}
  \caption{Performance comparison on Switchboard,
  trained on 300h.
  hybrid$^1$ is the IBM 2017 ResNet model \cite{saon2017ibmswb}.
  hybrid$^2$ trained with Lattice-free MMI \cite{hadian2018lfmmi}.
  CTC$^3$ is the Baidu 2014 DeepSpeech model \cite{hannun2014deepspeech}.
  Our attention model does not use any language model.}
  \label{tab:perfswb}
\end{table}

\section{Maximum expected BLEU training}

We implement expected risk minimization,
i.e.\ expected BLEU maximization or expected WER minimization,
following \cite{prabhavalkar2017minweratt,edunov2017maxbleu}.
The results are still preliminary but promising.
We do the approximation by beam search with beam size 4.
For a 4 layer encoder network model,
with forced alignment cross entropy training,
we get 30.3\% BLEU,
and when we use maximum expected BLEU training,
we get 31.1\% BLEU.

\section{Pretraining}
\label{sec:pretrain}

\crnn{} supports very generic and flexible pretraining which
iteratively starts with a small model and adds new layers in the process.
A similar pretraining scheme
for deep bidirectional LSTMs acoustic speech models
was presented earlier \cite{zeyer17:lstm}.
Here, we only study a layer-wise construction of the
deep bidirectional LSTM encoder network of an encoder-decoder-attention
model for translation on the WMT 2017 German$\to$English task.
Experimental results are presented in \Cref{tab:pretrain}.
The observations very clearly match our expectations,
that we can both greatly improve the overall performance,
and we are able to train deeper models.
A minor benefit is faster training speed of the initial pretrain epochs.

\begin{table}[th]
\setlength{\tabcolsep}{0.3em}
\centerline{
\begin{tabular}{|c|c|c|}
\hline
  encoder &  \multicolumn{2}{c|}{BLEU [\%]} \\
  num. layers & no pretrain & with pretrain \\
  \hline \hline
  2 & 29.3 & - \\
  3 & 29.9 & - \\
  4 & 29.1 & 30.3 \\
  5 & - & 30.3 \\
  6 & - & 30.6 \\
  7 & - & \textbf{30.9} \\
 \hline
\end{tabular}}
  \caption{Pretraining comparison.
  }
  \label{tab:pretrain}
\end{table}  

In preliminary recurrent attention experiments for speech recognition,
pretraining seems very essential to get good performance.

Also, we use in all cases a learning rate scheduling scheme,
which lowers the learning rate if the cross validation score does not improve enough.
Without pretraining and a 2 layer encoder in the same setting as above,
with a fixed learning rate, we get 28.4\% BLEU, where-as with learning rate scheduling,
we get 29.3\% BLEU.

\section{\crnn{} features}
\label{sec:features}

Besides the fast speed, and the many features such as pretraining,
scheduled sampling \cite{bengio2015schedsmpl},
label smoothing \cite{szegedy2016labelsmth},
and the ability to train state-of-the-art models,
one of the greatest strengths of \crnn{} is its flexibility.
The definition of the recurrent dependencies
and the whole model architecture are provided in a very
explicit way via a config file.
Thus, e.g.\ trying out a new kind of attention scheme,
adding a new latent variable to the search space,
or drastically changing the whole architecture,
is all supported already and does not need any more
implementation in \crnn{}.
All that can be expressed by the neural network definition
in the config. A (simplified) example of a network definition is given
in Listing \ref{lst:config}.

\begin{lstlisting}[float=*,floatplacement=tbp,caption={\crnn{} config example for an attention model},
label={lst:config},captionpos=b,
frame=single,basicstyle=\ttfamily\scriptsize,language=Python,
stringstyle=\color{deepgreen},commentstyle=\color{gray}]
network = {
# recurrent bidirectional encoder:
"src": {"class": "linear", "n_out": 620}, # embedding
"enc0_fw": {"class": "rec", "unit": "nativelstm2", "n_out": 1000, "direction": 1, "from": ["src"]},
"enc0_bw": {"class": "rec", "unit": "nativelstm2", "n_out": 1000, "direction": -1, "from": ["src"]},
# ... more encoder LSTM layers

"encoder": {"class": "copy", "from": ["enc5_fw", "enc5_bw"]},
"enc_ctx": {"class": "linear", "from": ["encoder"], "n_out": 1000},

# recurrent decoder:
"output": {"class": "rec", "from": [], "unit": {
  "output": {"class": "choice", "from": ["output_prob"]},
  "trg": {"class": "linear", "from": ["output"], "n_out": 620, "initial_output": 0},
  "weight_feedback": {"class": "linear", "from": ["prev:accum_a"], "n_out": 1000},
  "s_tr": {"class": "linear", "from": ["s"], "n_out": 1000},
  "e_in": {"class": "combine", "kind": "add", "from": ["base:enc_ctx", "weight_feedback", "s_tr"]},
  "e_tanh": {"class": "activation", "activation": "tanh", "from": ["e_in"]},
  "e": {"class": "linear", "from": ["e_tanh"], "n_out": 1},
  "a": {"class": "softmax_over_spatial", "from": ["e"]},
  "accum_a": {"class": "combine", "kind": "add", "from": ["prev:accum_a", "a"]},
  "att": {"class": "generic_attention", "weights": "a", "base": "base:encoder"},
  "s": {"class": "rnn_cell", "unit": "LSTMBlock", "from": ["prev:trg", "prev:att"], "n_out": 1000},
  "readout": {"class": "linear", "activation": "relu", "from": ["s", "prev:trg", "att"], "n_out": 1000},
  "output_prob": {"class": "softmax", "from": ["readout"], "dropout": 0.3, "loss": "ce", 
  	          "loss_opts": {"label_smoothing": 0.1}}
}},
"decision": {"class": "decide", "from": ["output"], "loss": "bleu"}
}
\end{lstlisting}

Each layer in this definition does some computation,
specified via the \texttt{class} attribute, and gets its input from other layers
via the \texttt{from} attribute, or from the input data, in case of layer \texttt{src}.
The \texttt{output} layer defines a whole subnetwork,
which can make use of recurrent dependencies via a \texttt{prev:} prefix.
Depending on whether training or decoding is done, the \texttt{choice} layer class
would return the true labels or the predicted labels.
In case of scheduled sampling
or max BLEU training,
we can also use the predicted label during training.
Depending on this configuration, during compilation of the computation graph,
\crnn{} figures out that certain calculations can be moved out of the recurrent loop.
This automatic optimization also adds to the speedup.
This flexibility and ease of trying out new architectures and models
allow for a very efficient development / research feedback loop.
Fast, consistent and robust feedback greatly helps the productivity and quality.
This is very different to other toolkits which only support
a predefined set of architectures.

To summarize the features of \crnn{}:
\begin{itemize}
\item flexibility (see above),
\item generality, wide range of models and applications,
such as hybrid acoustic speech models, language models
and attention models for translation and speech recognition,
\item fast CUDA LSTM kernels,
\item attention models, generic recurrent layer, fast beam search decoder,
\item sequence training (min WER, max BLEU),
\item label smoothing, scheduled sampling,
\item TensorFlow backend and the old Theano backend,
which has a separate fast attention implementation \cite{doetsch2016:bidir-dec-att},
fast CUDA MDLSTM kernels \cite{voigtlaender16:mdlstm},
as well as fast sequence training \cite{zeyer17:fasterseqtrain}.
\end{itemize}

One feature which is currently work-in-progress is the support for self-attention
in the recurrent layer. The reason this needs some more work is because
we currently only support access to the previous time step (\texttt{prev:})
but not to the whole past, which is needed for self-attention.
That is why we did not present any Transformer \cite{vaswani2017ffatt} comparisons yet.

\section{Conclusion}

We have demonstrated many promising features of \crnn{}
and presented state-of-the-art systems in translation and speech recognition.
We argue that it is a convenient testbed for research and applications.
We introduced pretraining for recurrent attention models
and showed its advantages while not having any disadvantages.
Maximum expected BLEU training seems to be promising.

{
\centering
\begin{minipage}{1.0\linewidth}
\section*{Acknowledgments}
\begingroup
\setlength{\columnsep}{5pt}%
\setlength{\intextsep}{0pt}%
\begin{wrapfigure}{l}[5pt]{0.5\textwidth}
\includegraphics[width=0.5\textwidth]{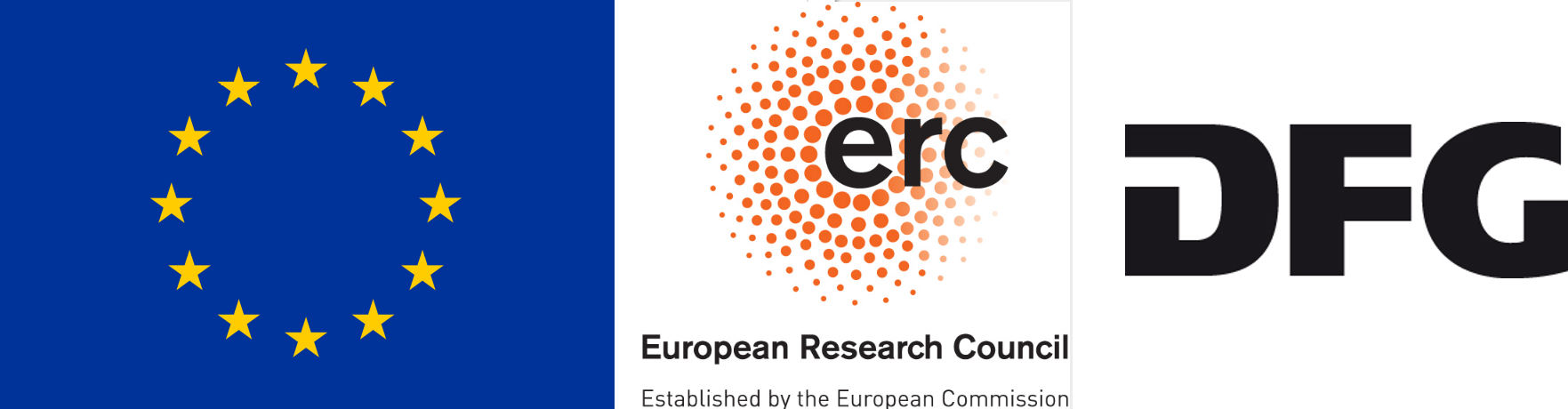}
\end{wrapfigure}
This research
has received funding from the European Research Council (ERC)
(under the European Union's Horizon 2020 research and innovation
programme, grant agreement No 694537, project "SEQCLAS") and
the Deutsche Forschungsgemeinschaft (DFG; grant agreement NE 572/8-1,
project "CoreTec"). Tamer Alkhouli was partly funded by the 2016 Google PhD fellowship for North America,
Europe and the Middle East.
The work reflects only the authors' views and none of the funding parties
is responsible for any use that may be made of the information it contains.

\endgroup
\end{minipage}
\vspace{-0mm}
\bibliography{acl2018}
\bibliographystyle{acl_natbib}


\end{document}